\documentclass{article}

\usepackage{arxiv}

\usepackage[utf8]{inputenc} % allow utf-8 input
\usepackage[T1]{fontenc}    % use 8-bit T1 fonts
\usepackage{hyperref}       % hyperlinks
\usepackage{url}            % simple URL typesetting
\usepackage{booktabs}       % professional-quality tables
\usepackage{amsfonts}       % blackboard math symbols
\usepackage{nicefrac}       % compact symbols for 1/2, etc.
\usepackage{microtype}      % microtypography
\usepackage{lipsum}
\usepackage{makeidx}
\usepackage{graphicx}
\usepackage{amsmath,amssymb} % define this before the line numbering.
\usepackage{color}

\usepackage{microtype}
\usepackage{graphicx}
\usepackage{subfigure}
\usepackage{booktabs} % for professional tables
\usepackage{amsmath}
\usepackage{amssymb}
\usepackage{subfigure}
\usepackage{multirow}
\usepackage{xcolor}

\def\ie{\emph{i.e.~}}
\def\eg{\emph{e.g.~}}

\title{Supervised Deep Kriging for Single-Image Super-Resolution}

\author{Gianni FRANCHI \\
Institute for Vision and Graphics, University of Siegen, Germany\\
\texttt{gianni.franchi@uni-siegen.de}
\And
Angela YAO\\
University of Bonn, Germany\\
\texttt{yao@cs.uni-bonn.de} 
\And
Andreas KOLB\\
Institute for Vision and Graphics, University of Siegen, Germany\\
\texttt{andreas.kolb@uni-siegen.de}}
\begin{document}
\maketitle

\begin{abstract}
We propose a novel single-image super-resolution approach based on the geostatistical method of kriging. Kriging is a zero-bias minimum-variance estimator that performs spatial interpolation based on a weighted average of known observations.  Rather than solving for the kriging weights via the traditional method of inverting covariance matrices, we propose a supervised form in which we learn a deep network to generate said weights.  We combine the kriging weight generation and kriging process into a joint network that can be learned end-to-end.  Our network achieves competitive super-resolution results as other state-of-the-art methods.  In addition, since the super-resolution process follows a known statistical framework, we are able to estimate bias and variance, something which is rarely possible for other deep networks.
\end{abstract}

% keywords can be removed

\section{Introduction}

Super-resolution aims to transform low resolution (LR) images into images with high resolution (HR). In computer vision, super-resolution is relevant for applications where high-frequency information and detailing is desirable yet not always captured, \eg medical imaging, satellite imaging, surveillance, etc. Our interest is in single image super-resolution (SISR), a special case where only one LR image is available. SISR is an ill-posed inverse problem with multiple solutions, \ie multiple HR images could lead to the same LR image after applying a low-resolution filter.

% Super-resolution has
SISR been solved with many different approaches.  Early approaches relied on spatial relationships present in the image and include nearest neighbours~\cite{chang2004super}, bi-cubic interpolation~\cite{keys1981cubic}. %and and the kriging methods~\cite{matheron1962traite,cressie2015statistics}.~\AY{I would prefer not to introduce kriging methods here; first, most people in the community do not know it, secondly it takes away from the uniqueness of your presentation here}.  
Other techniques build upon the assumption that the underlying signal is sparse, \eg compressive sensing~\cite{donoho2006compressed} and dictionary learning~\cite{yang2012coupled}. These early approaches tend to yield HR images which are blurry and or noisy, due mainly to the fact that these methods are crafted heuristically.  As a result, they cannot sufficiently recover enough information to yield sharp photo-realistic images.

Deep learning, and more specifically, convolutional neural networks (CNNs) \cite{lecun1998gradient}, has afforded significant performance gains in SISR.  The newest frameworks ~\cite{kim2016accurate,kim2016deeply,lim2017enhanced,taiimage,Sajjadi_2017_ICCV} learn feature representations for super-resolution in a supervised end-to-end manner and are achieving ever-improving performance both in terms of speed and quantitative evaluation.  However, despite their impressive results, most deep networks are being used as black boxes to make point estimates. The regressed outputs represent only a mean value and give no indication of model uncertainty. In the case of SISR, this means that we make (possibly very accurate) guesses for the unknown pixels, but we have no idea how good these guesses are.  This can be highly problematic for applications such as microscopy or medical imaging, where model confidence is just as important as the actual output.  

% we do not know for each resolve pixel what performances we should expect. Using a testing set and under the assumption that the image has the same distribution as the one of the testing set we can have some performance estimation. However, this performance estimate is the same for all the pixels of all the images. It happens that a DL framework might be good with some textural information and bad with others. Having performance estimate that is more accurate is a real need. This major issue might be problematic, for example in microscopy or medical images being sure that a given pixel is correct, is more important than applying super-resolution algorithm.
%one drawback from these networks remains: we cannot have access to uncertainty.~\AY{variance is not estimated, refer to work by Yarin Gal and other Bayesian deep learning approaches~\cite{gal2016uncertainty}} Concretely that means that when we super resolve an image 
 
We would like to leverage the strong learning capabilities of deep networks and embed them within a known statistical framework for which we can derive model uncertainty. As such, we propose  \emph{deep kriging} to solve SISR.  Kriging~\cite{MatheronBook,cressie2015statistics} is a geostatistics method used for spatial interpolation of attributes such as topography and natural resources. It has close relations to Gaussian process (GP) regression, although some sources oversimplify and describe the two as being equivalent. %Both kriging and GP regression consider treats the data as a random field  as a random field that the data is a random field : a random variable that depends on the space. 
The main difference is that kriging makes stationary and ergodic assumptions on the random field representing the data, while GP regression considers it as a Gaussian process.\footnote{We refer the reader to our supplementary materials for a more detailed comparison of the two.} 

% considers that the random field is stationary and ergodic contrary to GP regression that consider that process is a Gaussian Process.
%~\AY{add paragraph explaining relationship of kriging to Gaussian process regression; reference paper on gaussian process regression for superresolution and differences between our approach and theirs}
% Link with Gaussian process regression has been established in [ref].

% able to interpolate an image by learning the spatial relation present on the image. A more detailed introduction to kriging is written in section 3.1.  
% To solve this issue we propose to work with a hybrid between a DL framework and a statistical framework. The DL framework used is a deep dynamic local residual network. We use a deep dynamic local network ~\cite{de2016dynamic}, that consists of a network that learns a local filter that would be applied to an image. In addition, our network is based on a residual network since this kind of network has proved to achieve good results. The statistical framework used is the kriging. 

Kriging interpolates unknown values by taking a weighted average of known values and results in an unbiased estimator with minimum variance.
%~\AY{add one sentence about the properties of the method?} 
One can solve for the weights by inverting a covariance matrix computed on the known instances.  Kriging makes for a natural extension to the application of SISR, where HR images are interpolated from pixels observed in the LR version of the image. The computational cost of the kriging is cubic with respect to the number of known instances; hence it is preferable to work locally. However, interpolated results are often overly smooth and depend on the choice of the covariance function.%~\AY{do we have a citation to support this?}. 

Our contributions can be summarized as follows:

\begin{itemize}
\item We propose a novel deep learning-based statistical estimator with SISR results competitive with state-of-the-art.
\item We propose a deep kriging approach that solves for kriging weights and performs the kriging in a single network; this allows us to perform a supervised form of spatial interpolation which can be applied not only to SISR, but other geostatistical computations such as field reconstructions.
\item Our proposed deep kriging is a hybrid deep learning and statistical framework for which we can derive statistical bias and variance; such measures of model performance and uncertainty are not commonly available for other deep networks, making us the first to model and compute pixel uncertainty in deep SISR approaches.  
\end{itemize}

%-------------------------------------------------------------------------

\section{Related work}

%% P2: deep learning superresolution approaches

%% P3: use of Kriging (and Gaussian process regression) in computer vision; 

% Recent works performing single image super-resolution are mostly based on deep neural networks \cite{kim2016accurate,kim2016deeply,lim2017enhanced,taiimage,Sajjadi_2017_ICCV}. Our technique is comparable to these. However, we combine a deep neural network with a kriging regression in order to have a better understanding of what we learned. This kriging interpolation is performed thanks to a deep local dynamic filter~\cite{de2016dynamic}. On this section, we will focus on the deep learning framework performing SSIR, on other Kriging or Gaussian Process Regressions (GPR) that have been combined with deep learning framework, and also in deep local filters.
% \subsection{SISR}
Prior to the use of deep learning, SISR approaches applied variants of dictionary learning~\cite{freeman2002example,donoho2006compressed,yang2012coupled,peleg2014statistical,dong2011image}. 
Patches were extracted from the low-resolution images and mapped %. Thanks to a learning procedure these patches were mapped 
to their corresponding high-resolution version which are then stitched together to increase the image resolution.  Other learning-based approaches to increase image resolution include %Later, different engineer feature machine learning algorithm 
~\cite{zhao2003wavelet,zhang2012single,anbarjafari2010image}. % were proposed to increase the resolution of images. %Among them one can note the use of shallow networks [ref].%~\AY{this last sentence is way too vague -  it describes every single non-supervised SR technique??!}

State-of-the-art SISR methods are deep-learning-based~\cite{lim2017enhanced,kim2016accurate,kim2016deeply,Sajjadi_2017_ICCV,taiimage}.  The VDSR~\cite{kim2016accurate} and DRCN~\cite{kim2016deeply} approaches showed the benefits of working with image residuals for super-resolution. % In particular, this allows networks to focus on learning only the missing high-frequency information.
The DRRN approach~\cite{taiimage} generalizes VDSR and concludes that the deeper the network, the better the super-resolved image. We also use a residual network in our approach, but unlike all the other deep SISR methods, we are solving for a set of filter weights to perform the super-resolution with our network rather than directly estimating the HR image itself.

Several unsupervised techniques have also been developed for SISR~\cite{xu2010two,marquina2008image}, though their performance is usually poorer than supervised approaches.  One work of note~\cite{he2011single}, uses GP regression to perform super-resolution and resembles ours in spirit in that we both model pixel intensities as a random process regressed from neighbouring pixels.  However,~\cite{he2011single} is  unsupervised while our approach learn the weights from supervise examples.

Our proposed approach can be thought of as a form of local filtering~\cite{de2016dynamic}, and is most similar in spirit to works which combine deep learning and GP regression~\cite{wilson2016deep,damianou2013deep}.  However, we differ from these techniques since we do not apply GP regression but a modified version of local kriging.  These techniques are similar to us in that they also consider the data to follow a random process. %, and also in the form of the solution.
%Wilson~\etal~\cite{wilson2016deep} propose to learn the parameters of a covariance matrix with a a deep Bayesian framework.  In~\cite{damianou2013deep}, Damianou~\etal concatenate GPs, where the output of one GP is used by another process as a spatial distribution.  
However, we do not explicitly learn the covariance matrix, which offers us more flexibility in the relationships we wish to express. Furthermore, we treat each image as following a random process; the set of training images is then a set of random processes, while the aforementioned works consider all the training data to follow the same process.

\section{Deep Kriging}
We begin with a short overview on classical kriging  in Section~\ref{sec:classicK} before we introduce our proposed method of supervised kriging in Section~\ref{sec:deepkrig} and elaborate on its statistical properties in Section~\ref{sec:stats}.  Finally, we show how the proposed form of deep kriging can be implemented in a deep network in Section~\ref{sec:implementation}. 

\subsection{Classical kriging}~\label{sec:classicK}

Consider an image $f$ as a realization of a random field $F$. This random field is a collection of random variables, \ie $F = \{ F_i \}$ where %$i$ stands for a spatial positions and 
each $F_i$ is a random variable and $i$ is a position index. Unknown values on $f$ at position $x^*$, $\hat{f}_*$\footnote{For concise notation, we use $\hat{f}_*$ to denote $\widehat{f}(x^*)$, $f_i$ to denote $f(x_i)$ and $w_i^* =w_i(x^*)$.}, can be interpolated linearly from $n$ known realizations $f_i$, with normalized weights $w_i$, \ie
\begin{equation}
% \widehat{f}(x^*)=\sum_{i=1}^n \omega_i(x^*)f(x_i), \quad \text{with }\sum_{i=1}^n \omega_i(x^*) = 1.
\hat{f}_*=\sum_{i=1}^n \omega_i(x^*)f_i, \quad \text{with }\sum_{i=1}^n \omega_i(x^*) = 1.
\end{equation}
In classical kriging, all known realizations are used in the interpolation and $n$ is the number of pixels in the image, though local variants have been proposed and studied in \cite{pronzato2017bayesian,meier2014local}. 

The weights $\{\omega_i\}$ are found by minimizing the true risk: 

% \begin{eqnarray}\nonumber
\begin{equation}\label{eq:truerisk}
R(\omega) =  \mathbb{E}\left[(F_* - \hat{F}_*)^2\right]\\
% \\                         &=&
=
\mbox{var}\left(
                       %f(x^*)-\sum_{i=1}^n \omega_i(x^*) f(x_i)
F_*-
% \sum_{i=1}^n \omega_i(x^*) f_i               
\hat{F}_*
                         \right).
% \end{eqnarray}
\end{equation}
%\AY{%why do we want to introduce it starting from the true risk?  we do not have access to f(x*)? 
%Should the correct form be $\mathbb{E}\left[(f_* - \hat{f}_*)^2\right]$?}~\correction{ok}

\noindent We can equate the $\mathbf{E}[\cdot]$ with the $\text{var}(\cdot)$ term in Eq.~\ref{eq:truerisk} because the constraint that the weights must sum up to $1$ implies that %$\mathbb{E}\left( f(x^*)-\sum_{i=1}^n w_i(x^*) f(x_i)\right) = 0$.
$\mathbb{E}[ f_*-\hat{f}_*]\!\!=\!\!0$.  However, since we do not have access to different realizations of the random field, this variance cannot be solved directly. %  and consequently, this variance is an unknown parameter. 

To infer the variance from a single event, the theory of geostatistics replaces the classical statistics assumption of having independent and identically distributed (iid) random variables with assumptions of stationarity and ergodicity on the random field. The random field is assumed to be first and second order stationary. This means that $\mathbb{E}_{F^i} (f^i(x))$ does not depend on $x$ and that $\mathbb{E}_{F^i} (f^i(x)\times f^i(x+\tau))$ depends only on $\tau$. In addition, the field is assumed to be second order ergodic, so a covariance or mean estimate in the probability domain is equivalent inthe spatial domain. This second assumption implies that an empirical covariance
%~\AY{we jumped from variance to covariance}~\correction{we need the covariance after}
can be estimated and depends only on a distance $\tau$ between two realizations, \ie $\tau = \|x_i-x_j\|$. 
However, it is difficult to work directly with the empirical covariance, since it can be noisy and also is not guaranteed to form a positive semi-definite matrix. % is not necessarilynegative definite,  then we might not have a global minimum. Secondly, we aim for results which can generalize and do not overfit to the training data.~\AY{so what?  I don't see how this point follows from what you said about the empirical covariance.}
%~\AY{how much of this is specific to the case of Kriging and how much of this is standard in statistics / machine learning?  Don't we almost always work with the empirical covariance and or empirical loss?}~\correction{Here we work with the true covariance and not the empirical one}
%To overcome the drawbacks of working directly with the empirical covariance, 
As such, in kriging, we use the true covariance by fitting a parametric model. While different models can be used, the most common one is Gaussian, with covariance $C_{ij}$ between points $x_i$ and $x_j$ defined as 
$
C_{ij} = C_0 \exp{\left(-\tfrac {1}{\sigma^2}\|x_i-x_j\|^2\right)},
$ where $C_0$ and $\sigma$ are parameters of the model.

A Lagrange multiplier (with constant $\lambda$) can be used to minimize the risk function with the constraints on the weights $w_i$, leading to the following cost function:
\begin{eqnarray}~\label{KrigingCostMu}
\mathcal{L}(w)  = \text{var}\left( F_*-\hat{F}_*\right) + 2\lambda \left( \sum_{i=1}^n w_i(x^*)-1 \right),
% & = \sum_{i,j=1}^n w_j(x^*)w_i(x^*)C_{ji} - 2 \sum_{i=1}^n w_i(x^*)C_{*i} +C_{**} \nonumber + 2\lambda \left( \sum_{i=1}^n w_i(x^*)-1 \right). \nonumber
\end{eqnarray}
and the associated solution expressed in matrix form as:
%and equalize to zero, and we obtain the following system:
% \begin{align}
%  \begin{cases} % \left\{
%     % \begin{array}{l}
%       2\sum_{j=1}^n  w_j(x^*)C_{ji}-2 C_{*i} - 2 \lambda  =0 \,\,\, \forall i \in [i,n] \\ 
%       \sum_{j=1}^N w_j(x^*)=1.
%     % \end{array}
%     \end{cases}
% % \right.
% \end{align}
% \footnotesize
% \begin{eqnarray}
%  \left\{
%     \begin{array}{ll}
%         2\sum_{j=1}^n  w_j(x^*)C(x_j,x_i)-2 C(x^*,x_i) - 2 \mu   =0 & \forall i \in [i,n] \\ \\
%       \sum_{j=1}^N w_j(x^*)=1                                      &~
%     \end{array}
% \right.
% \end{eqnarray}
% \normalsize
%\noindent ~\AY{move either equation system or matrix to appendix if need space; redundant}. We can also express the above system in matrix form as: 
{\small
\begin{eqnarray}~\label{eq:covmat}
\begin{pmatrix} 
% C(x_1-x_1) & \ldots & C(x_1-x_n)&1 \\
C_{11} & \ldots & C_{1n} &1 \\
\vdots & \ddots & \vdots & \vdots \\
% C(x_1-x_n) & \ldots & C(x_n-x_n)&1 \\
C_{1n} & \ldots & C_{nn}&1 \\
1 & \ldots & 1 & 0
\end{pmatrix}
\begin{pmatrix} 
w_1 (x^*)\\
\vdots \\
w_n(x^*) \\
-\lambda
\end{pmatrix} = 
\begin{pmatrix} 
% C(x_1-x^*)\\
C_{1*}\\
\vdots \\
% C(x_n-x^*)\\
C_{n*}\\
1
\end{pmatrix}.
\end{eqnarray}
}
\normalsize
The process of kriging then is reduced to solving for the weights $w_i$.  This involves inverting the covariance matrix on the left.  Since the matrix is of dimension $(n+1)$, where $n$ is the number of observed pixels in the image (patch), it can be very computationally expensive. Even though one can limit $n$ with local kriging, $n$ needs to be sufficiently large in order to accurately capture and represent the spatial relationships.  We aim to bypass the matrix inversion and still maintain sufficient generalization power by using a deep network to directly learn the weights instead.
% To avoid the matrix inversion, we propose using a neural network to directly learn the weights instead. 
% ~\AY{1-2 sentences about striking a trade-off between computational expense of inverting the covariance matrix, yet be large enough in spatial coverage to ensure that the covariance matrix is sufficiently accurate; using supervised kriging in a deep network, we can strike this balance.}

%As such, we propose using a neural network to learn how to solve for the weights instead. This avoids the matrix inversion, but more importantly, %saves us from having to invert the $(n+1)$ matrix, but more importantly, 
%prevents us from having to depend on a fixed model contained within the one image.  
% The covariance matrix of classical kriging is a reproducing kernel. So we project the spatial position into an infinite dimension space. This is analogous to using a neural network with one hidden layer. It is well known in machine learning that when we use this kind of network, based on the universal approximation theorem~\AY{add ref from 6.4.1 deep learning text}, we can approximate any measurable function up to any desired amount of error.  However, it has been proven empirically that deep neural network performs better by learning the hidden structure by having multiple hidden layers. That is why we propose to use a neural network to learn the weights and so the kernel.  In some sense, the network learns the kernel matrix that needs to be inverted, such that we do not depend on a specific model.

\subsection{Supervised kriging}~\label{sec:deepkrig}
Consider $E$, a subset of the discrete space $\mathbb{Z}^2$, as a representation of the support space of a 2D image. We denote an image as $f \in \mathbb{R}$ and $f_i$ as its value at pixel $x_i \in E$.  This assumes we work with greyscale images, which is a standard assumption made in most SISR algorithms~\cite{kim2016deeply,kim2016accurate,taiimage}.% as most improvements come from the spatial information contribution of images and not the spectral one.}
% Our model assumptions are the following. We consider that we have a set of images called the training set that is composed of $n_1 \in \mathbb{N}$ pairs of images $( \tilde{f}_i, f_i )$ with  $ i \in [ 1,n_1 ]  $, where $\tilde{f}_i$ is a low resolution version of $f_i$ that has the same size of $f_i$.

% We assume that we are given a set of $n_1$ training image pairs $\{ (\tilde{f}^i, f^i)\}, i\in [ 1,n_1 ]$, where $\tilde{f}^i$ is an up-sampled low resolution version of $f^i$ with the same size as $f^i$.  We further assume that  $\tilde{f}^i$ follows a first and second moment stationary random process $\tilde{F}^i$. Let us write, the training set $\{ \tilde{f}^i, f^i \}_{i=1}^{n_1} \sim \mathcal{M}$, where $\mathcal{M}$ is a joint meta-distribution where each training pair  $(\tilde{f}^i, f^i) \sim \tilde{F^i} \times F^i $ follows its own random process.  
We assume that we are given a set of $n_1$ training image pairs $\{ (\tilde{f}^i, f^i)\}, i\in [ 1,n_1 ]$, where $\tilde{f}^i$ is an up-sampled low resolution version of $f^i$ with the same size. Our objective is to super-resolve the low-resolution images of the test set $\mathcal{N}_{\mbox{test}}= \{  \tilde{f}^i \}_{i=1}^{n_2} $.  We further assume that $\tilde{f}^i$ is a realization of a first and second moment stationary random process $\tilde{F}^i$. For convenience we denote $\tilde{F}^i$ the distribution of the random process. In addition $f$ is a realization of a random field $F$. We denote the training set as $\{ \tilde{f}^i, f^i \}_{i=1}^{n_1} \sim \mathcal{M}$, where $\mathcal{M}$ is a joint meta-distribution where each training pair  $(\tilde{f}^i, f^i) \sim \tilde{F^i} \times F^i $ follows its own random process.  

% We consider that the training set $\mathcal{N}_{\mbox{training}}= \{ \tilde{f}_i, f_i \}_{i=1}^{n_1} \sim \mathcal{M}$ where each subset $(\tilde{f}_i, f_i)  \sim \tilde{F_i} \times F_i $  for $i\in [1,n_1]$. $\mathcal{M}$ is a joint meta distribution where each pear of image follow it own random process. 
% 
%~\AY{move training and testing set information to later?}

In classical kriging, the estimated $\hat{f}(x)$ is expressed as a linear combination of all the observed samples \ie pixels of $f$. In the case of super-resolution, the observations come from the low-resolution image $\tilde{f}(x)$. Furthermore, we estimate $\hat{f}(x)$ as a linear combination of only local observations of some window radius $\mathcal{K} $, leading to the estimator:%~\AY{unused index; use window as a parameter rather than explicitly writing 3 and 5, also use only one variable rather than 2.} % local  in order to be able to handle huge images we work with a local version using sliding window of size $5 \times 5$. So the estimator we use is:

%the estimated $\hat{f}(x)$ is expressed as a linear combination of all the observed samples \ie pixels of $f(x)$, in our model, we consider that we do not observe $f(x)$, but $\tilde{f}(x)$. Furthermore in order to be able to handle huge images we work with a local version using sliding window of size $5 \times 5$. So the estimator we use is:

%~\AY{why do we change to $a$ here and not stick to $w$?  also it seems you change the notation of $w$ now to express the network parameters? consistency in terminology and wording, i.e. loss, $R$, cost function, risk, etc.}

\begin{equation}~\label{method:form1}
    \hat{f}^i_* = \mkern-24mu \sum_{k \in \{\|x_k-x^* \|_1 \leq \mathcal{K} \}}\mkern-40mu\omega_{k}(x^*) \tilde{f}_{k}^i \quad
\text{with} \mkern-24mu \sum_{k \in \{\|x_k-x^* \|_1 \leq \mathcal{K} \}}
\mkern-40mu \omega_{k}(x^*) = 1.
\end{equation}

% \begin{multline}
% \hat{f}^i_* = \sum_{k_1\subset \{k|\mbox{ }\|x_k-x^* \|_1 \leq \mathcal{K} \} }\omega_{k_1}(x^*) \tilde{f}_{k_1}^i \\
% \mbox{ with } \sum_{k_1\subset \{k|\mbox{ }\|x_k-x^* \|_1 \leq \mathcal{K} \}}\omega_{k_1}(x^*) = 1,
% \end{multline}
% it happens if the size of the sliding windows is $\mathcal{K}  =2$ then there are 25 weights $\omega_k$ such that $\|x_k-x^* \|_1 \leq 2$, which are the 25 nearest neighbors of $x^*$.

When no confusion is possible we denote $\hat{F}_*^i = \sum_{k_1=1}^{n}\omega_{k}(x^*)  \tilde{F}_{k}^i$, with $n = (2\mathcal{K}+1)^2$. We  found that a window  of $7\times7$ provided the best results. To find $\omega$, we want to minimize the true risk as given in Eq.~\ref{eq:truerisk}, leading to:
\begin{equation}
\label{method:form2}
R(\omega) =\sum_{i=1}^{n_1}\mathbb{E}_{F^i\times\tilde{F}^i}\left[\left(F_*^i- \hat{F}^i\right)^2\right].
\end{equation}
This risk is a compromise between the geostatistical model and the deep learning model, where the covariance of each field is learned through supervised learning.  Furthermore, if we replace the true expectation with the empirical one and the random field with its realization, we arrive at 
\begin{equation}
\label{method:form3}
R(\omega) =\sum_{i=1}^{n_1}\sum_{j=1}^{N_i}\left[\left(f_*^i-\sum_{k=1}^n \omega_{k}(x^*) \tilde{f}_{k}^i\right)^2\right],
\end{equation}
where $N_i$ is the number of pixels in image $f^i$.

Now, rather than solving for the weights $\omega$ by inverting the covariance matrix in Eq.~\ref{eq:covmat}, we propose learning this set of weights directly with a CNN. This allow us to capture and model more complex spatial relationships between the data than the classical model with covariances.  As such, the estimator becomes $\hat{f}(x, \tilde{f}, g, \theta)$, where $w = g(\tilde{f}, \theta)$, with $g$ and $\theta$ representing the CNN network function and its parameters respectively. 

% which is based on convolutional layers and is able to model more complex spatial relations between data, than the one model by classical covariances.  Hence we have  $a(x) = g(\tilde{f},\theta)$ where $\theta$
% represents the set of parameters used to learn $a(x)$ and $g$ the neural network function used. The true estimator should be written : $\hat{f}(x,\theta, g, \tilde{f})$, for simplicity we use the following simplified notation: $\hat{f}(x)$ and $a(x)$. 

\subsection{Statistical properties}~\label{sec:stats}
We can derive statistical properties for our proposed estimator $\hat{F}$. Given $\{ \omega_1, \ldots,\omega_n  \}$, the bias can be expressed as  
% When one builds an estimator one can wonder if it is consistent that means that it goes closer and closer to the true value  as  the number of samples increases.
% In this case, the bias of the estimator is equal to zero. The bias of the estimator we propose is:
% \begin{equation}
\small
\begin{align*}
\label{Stat:form1}
\text{bias}(x) & = \mathbb{E}_{\tilde{F}| \omega_1, \ldots,\omega_n} \big[\tilde{F}(x)-\hat{F}(x)  \big] \\
& = \mathbb{E}_{\tilde{F}| \omega_1, \ldots,\omega_n} \big[ \tilde{F}(x) \big]  -\mathbb{E}_{\tilde{F}| \omega_1, \ldots,\omega_n} \big[\hat{F}(x) \big] \\
& = \mathbb{E}_{\tilde{F}| \omega_1, \ldots,\omega_n}\left[\tilde{F}(x)\right] - \sum_{k_1=1}^{n}\omega_{k_1}(x_j) \mathbb{E}_{\tilde{F}| \omega_1, \ldots,\omega_n}\left[\tilde{F}(x)\right].
\end{align*}
\normalsize

% \end{equation}

% which is equal to:
% \begin{equation}
% \label{Stat:form2}
% \text{bias}(x) = \mathbb{E}_{F} \left( f(x) \right)  -\mathbb{E}_{\tilde{F}} \left(\hat{f}(x) \right). 
% \end{equation}

\noindent Since $\sum_{k_1=1}^{n}\omega_{k_1}(x_j) =1$ and the random field is first-order stationary, we are left with zero bias according to the weights of the neural network, \ie $\text{bias}(x)=0$.
In other words, our network does not add bias to the field $\tilde{F}$.  As such, from $F$ to $\tilde{F}$, if the two fields have the same first moment, then no bias added to $F$\footnote{Of course, we cannot account for the low resolution process that produced $\tilde{F}$}. 
% Using the fact that the first moment of the random fields are stationary and by replacing $\hat{f}$ by its value we have:
% \begin{equation}\label{Stat:form3}
% \text{bias}(x) = \mathbb{E}_{\mathcal{F}}\left[f(x)\right]  - \sum_{k_1=1}^{25}a_{k_1}(x_j) \mathbb{E}_{\tilde{F}}\left[\tilde{f}(x)\right] .
% \end{equation}
% Since $\sum_{k_1=1}^{25}a_{k_1}(x_j) =1$  we have: 
% \begin{equation}
% \label{Stat:form3}
% \text{bias}(x) =  0 .
% \end{equation}
% So our filter has no bias. 
%This point is critical -- a zero bias implies that with an infinite dataset our estimator can perfectly super resolve the image.
This point is critical since in classical statistics a good estimator should have zero bias and minimal variance.
% In addition on can be interested by the variance of the estimator which is :

By definition, the variance of our estimator $\hat{F}$ is 
\begin{equation}\label{Stat:form4}
V_{\tilde{F}}(x) = \mathbb{E}_{\tilde{F}} \left[ \hat{F}(x)^2 \right] - \left(\mathbb{E}_{\tilde{F}} [\hat{F}(x)] \right)^2.
\end{equation}
Again, we estimate the covariance given $\{\omega_1, \ldots,\omega_n\}$.
Since we assume that $\tilde{F}$ is second-order stationary, the covariance of $\tilde{F}$ depends only on the distance between two points, \ie 
\begin{equation}
\mathbb{E}_{\tilde{F}}  \left[ \tilde{F}(x_{k})\tilde{F}(x_{k'})\right] = \tilde{C}(\|x_{k}-x_{k'}\|_2)- \mu^2.
\end{equation}
By setting $\mu=\mathbb{E}_{\tilde{F}} [\tilde{F}(x)] $, we arrive at:
%~\AY{fix 25 in neighbourhood}
\small
\begin{equation}
\label{Stat:form8}
V_{\tilde{F}| \omega_1, \ldots,\omega_n}(x) = \mkern-36mu \sum_{\substack{(k,k')\in [1,(2\times\mathcal{K}+1)^2]^2}} \mkern-36mu \omega_{k}(x)  \omega_{k'}(x) \tilde{C}(\|x_{k}-x_{k'}\|_2)
\end{equation}
\normalsize
where the covariance $\tilde{C}(\cdot)$ is estimated from the low resolution image $\tilde{f}$. The variance depends on the position $x$. % We provide a more detailed derivation on the calculus of the variance in the supplementary materials.
Note, however, since we directly estimate the weights $\omega(x^*)$ without making assumptions on the covariance function, we cannot compute the variance and an approximation is needed to estimate the covariance empirically from $\tilde{f}$. Eq.~\ref{Stat:form8} however is particularly interesting since it shows that the estimator variance at $x$ is directly proportional to the variance of $\tilde{f}$ and the different values of $\omega$ near $x$. The bigger these values are, the more uncertain the estimator is.%~\AY{need one more sentence here to state the "so-what".}

\subsection{Network Implementation and Training}~\label{sec:implementation}

% In this section we will present the technical aspects regarding the network and the training procedure. 
% Similarly to [ref] 

\begin{figure*}[tp!]
\centering
   \includegraphics[width=0.90\linewidth]{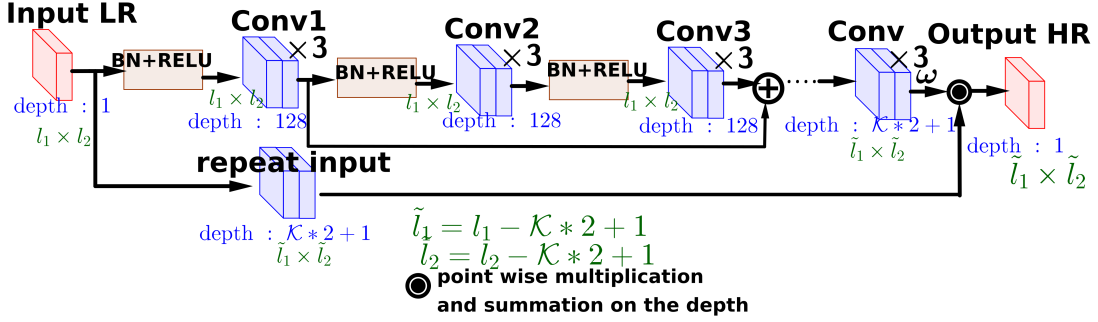}
	
\caption{Overview of the network structure. The input image goes through two branches. The first one calculates the weights $\omega$; the second copies the input image for an efficient application of the weights via point-wise multiplication.  
%and extracts the neighborhood of each pixel. Later, the weights $\omega$  are normalized and then applied to the neighborhood of each pixel. Finally, the results are summed over the neighborhood and which provide the output. 
}\label{fig:network}
\label{fig:multiple_shape1}
\end{figure*}

So far, the theory which we have presented on supervised kriging is generic and does not indicate how one should minimize the risk in Eq.~\ref{method:form3} to solve for the kriging weights.  We solve for the weights with a CNN network which we %This network can follow many possible architectures and we pick the same architecture as~\cite{taiimage}, 
show in Fig.~\ref{fig:network}.  The weight estimation branch is composed of a residual network of 9 residual units; the residual unit itself has 2 convolutional layers, each preceded by a batch normalization~\cite{ioffe2015batch} and a ReLU. 

 This branch follows a similar architecture as~\cite{taiimage,kim2016accurate}, with the difference that they apply this architecture to directly estimate the HR image, while we use it to estimate our kriging weights, \ie as a local dynamic filter~\cite{de2016dynamic}. In addition, contrary to \cite{taiimage} our network is not recurrent.  ~%\AY{please check that this section is now correct and matches your diagram -- I tried to remove the mention of recursion, etc.}

% Our network is composed of a series of residual units as illustrated in figure \ref{fig:RSU}. 
% A residual unit is a small network composed of 3 convolutions layer each of them is followed by a batch normalization to increase the generalization power and after the batch normalization of the two first convolutions, we apply a relu activation function. At the end of the last convolution layer, a batch normalization is applied and its output is sum with the first layer of the recursive block. 
% In our case since we consider just one block similarly to [ref] the signal added to each residual unit is the output of the first convolution called conv1 in figures \ref{fig:RSU} and \ref{fig:network}.
 
% Our network as illustrated in figure \ref{fig:network} is composed of a structure called first convolution (conv1) followed by a network 1 which is a recursive block composed of U residual units, followed by a prediction convolution. Contrary to [ref] each residual unit is composed of 3 convolutions layer instead of two in the article. The use of this residual units help to learn complex structures in the data, and avoid overfitting has proven in [ref] due to the fact that the network learns just the residues. 

% To reduce the number of parameters and thus avoid over fitting, all the residual unit are learned recursively similarly to [ref]. That means the same weights are used for different residual units.
The network learns a total of 20 convolution layers to determine the kriging weights.  The first 19 are of depth 128, the last prediction layer has depth $(2\times\mathcal{K}+1)^2$ and outputs the kriging weight vector $\omega(x)=[\omega_{1}(x),\ldots,\omega_{(2\times\mathcal{K}+1)^2}(x)]$.  The weights are applied to the repeated input image via a point-wise multiplication and summation along the depth dimension to yield the HR output. Overall, the network with $\mathcal{K} = 3$ is very lightweight, inference on a $320 \times 480$ sized image takes only $0.10$ seconds with a titan X. 

% with only $352k$ parameters; applying the network takes $0.25$ seconds on an image of $320 \times 480$ resolution.  
% We emphasize that this network is light since the output is just a linear combination of the input data.%~\AY{this is a very lightweight network! we should emphasize this!}

Note our network is actually learning how to estimate the kriging weights $\omega$, \ie as a local dynamic filter.   
% The prediction of $\omega$ is crucial since the network does not learn to estimate the output directly; rather it learns how to generate krigigna  , but a local dynamic filter to apply to the image. 
To derive this dynamic filter end-to-end, we adapt the following formulation of the Eq.~\ref{method:form1} in convolution terms:
%~\AY{consider placing the convolutional form directly in earlier equation}
\begin{equation}
\label{method:form443}
\tilde{f}_i(x^*) = \mkern-36mu \sum_{k \in \{\|x_k-x^* \|_1 \leq \mathcal{K} \} } \mkern-36mu \omega_{k}(x^*) \left( \tilde{f}_i\ast h_{k} \right)(x_{k}) \qquad \text{ with } \qquad \mkern-36mu \sum_{k \in \{\|x_k-x^* \|_1 \leq \mathcal{K} \} } \mkern-36mu \omega_{k}(x^*)=1
\end{equation}
with $h_{k}(x) = \delta(x-x_{k})$ or the Dirac function. The $h_{k}$ filter is denoted by \emph{``repeat input''} in Fig.~\ref{fig:network}, since applying the Dirac functions in each position $x^*$ is a way to extract the pixel neighbourhood at position  $x^*$. %The deep dynamic local filter [ref] which is based on convolution layers is able to model more complex spatial relationships between data. Contrary to [ref] that learns the dynamic filter on an image and applies it to another one we learn and apply the filter on the same image. Then the output of this filter is 25 activation maps where at each position $x$ a pixel of an activation map corresponds to a nearest neighbor of this position according to a specific Dirac function.
After having determined the weights, we normalize them so that they sum up to $1$, as given by the constraint in Eq.~\ref{method:form443}.  We point-wise multiply the normalized weights to the output of the repeat filter; to arrive at the super-resolved image, we simply sum along the depth.  

% To avoid border effects at the last convolution of the network and the repeat input filter, we do not pad the input.  As such, the output's spatial size is reduced spatially by $2\mathcal{K}$ in each dimension.  When the kriging weights are learned, the spatial support of the kriging actually extends beyond the local window to which the weights are applied.  

%~\AY{you define the depth below, based on the amount of recursion, but don't specify how the depth relates to the spatial window of the Kriging?}
%depends on various parameters among them the depth $d$ of the network. To define the depth of a fully connected network, one has to visual a fully connected network has a big filter that is applied to an image.
%Then one might be interested in the spatial size of the filter. The depth $d$ of the network corresponds to the distance from the center of the sliding windows to  its edge and is defined by :
% The efficiency and speed of a network depends on various parameters among them the depth $d$ of the network and its receptive field. %The depth of a network corresponds to the number of convolution layers while the receptive field consists in visualizing a fully connected network has a big filter that is applied to an image.

We implement our network in Tensorflow, minimizing the loss in Eq.~\ref{method:form3} with the Adam optimizer, a batch size of 8 and a learning rate of $10^{-4}$.  We apply gradient clipping and dropout between the last ReLU and the convolution layer that predicts $\omega$, with a dropout rate of 0.3.  The effective depth of our network is $d = 2+2\times U$, with $U$ the number of times the residual units are applied.  In our case, we use $U=9$. %The receptive field of our network is is defined by : $r = 2+8\times U + 2\times \mathcal{K}$.  For a window of $7 \times 7$, \ie $\mathcal{K} = 3$, this leads to a receptive field size of $80 \times 80$.

\section{Experiments}

\subsection{Datasets, Pre-processing and Evaluation}
For training, we follow~\cite{taiimage,kim2016accurate} and use the 291 images combined from~\cite{yang2010image} and the Berkeley Segmentation datasets~\cite{MartinFTM01} .  We further augment the data by rotating ($90^{\circ}$, $180^{\circ}$ and $270^{\circ}$) and scaling the data ($\times2$, $\times3$, $\times4$). We limit to these fixed grades of rotation and scaling to follow the same protocol as ~\cite{taiimage}.
%~\AY{why do we limit to only these fixed grades of rotation and scaling?}
%Like~\GF{ref}, since the spatial receptive field of our network is~\AY{X}, we work with patches of this size, sampled from the augmented dataset with a stride of $21$.
Like~\cite{taiimage,kim2016accurate}, we work with patches of size $31 \times 31$, sampled from the augmented dataset with a stride of $21$.
% Following the experimental protocol of proposed in [ref], we use as training set and validation set two data sets that we have combined. The first one [ref] is the composed of natural 91 images that contain much textural information. The second one is the Berkeley Segmentation Dataset, which is composed of 200 natural images and was originally set for segmentation. So the training set is composed of 291 images. Using the Matlab database implementation of [ref] we rotate the original images by $90°$,$180°$ and $270°$, and use different scales of the image $\times 2$,$\times 3$, and $\times 4$. In addition instead of working with the full image, we work with patches of size $31 \times 31$ with a stride of $21$ that has been noticed to improve the results in [ref]. In particular this last technique increase considerably the size of data set.
Testing is then performed on the four commonly used benchmarks: \emph{Set 5} \cite{bevilacqua2012low}, \emph{Set 14} \cite{kim2016deeply}, \emph{B100} \cite{MartinFTM01} and \emph{Urban 100} \cite{huang2015single}. To be consistent with~\cite{taiimage,kim2016accurate}, we convert the RGB images to YCbCr and only resolved the Y component.  This is then combined with an upsampled Cb and Cr component and converted back into RGB. Up-sampling is done via bi-cubic interpolation.

% \subsection{Evaluation Measures}
We evaluate the resulting super-resolved images quantitatively with the peak signal-to-noise ratio (PSNR):
$
\text{PSNR}(\tilde{f},f_{ \text{ref}}) = \log _{10}\left({\frac {255^{2}}{\sum _{i=1}^{N}(\tilde{f}(x_i)-f_{ \text{ref}}(x_i))^{2}}}\right),
$   
where $f_{ \text{ref}}$ and $\tilde{f}$ are the ground truth and the super-resolved images respectively and $N$ refers to the number of pixels in the (high-resolution) image.  A higher PSNR corresponds to better results. We also evaluate using Structural Similarity (SSIM)~\cite{wang2004image}, which measures the similarity between two images. The closer the SSIM value is to $1$, the more similar the super-resolved image to the ground truth high resolution image.

\subsection{Comparison to state-of-the-art}
We compare our PSNR and SSIM measures for the different scales against state-of-the-art in Table~\ref{table:results_Superresolution}.  The first two methods, bi-cubic and local kriging are unsupervised, while all others are supervised approaches.  We use the Matlab implementation of bicubic interpolation.  

For the local kriging, we use a neighbourhood of $90 \times 90$ and a stride of $81$.  At this sparse setting, unsupervised kriging does very well in comparison to bicubic interpolation. %As can be expected, the supervised kriging does much better and has 
However, it is already extremely slow, since for each patch, we need to compute the empirical covariance, true covariance, and then invert the covariance matrix. In total, it takes approximately 1 minute for an image of size $320 \times 480$. In comparison, our proposed deep kriging, on the same image, applied densely is one magnitude faster, and takes only $0.25$ seconds with a titan x.  % Furthermore, as expected, the supervised kriging has a much higher PSNR and SSIM.

% As one can see from Table \ref{table:results_Superresolution}, that  supervised kriging does a better job at super-resolution than local unsupervised kriging. This is due to the fact that the network learns a complex covariance matrix adapted to each image. For the unsupervised kriging, we use  

Looking at our approach (reported in the second last column in Table~\ref{table:results_Superresolution}) with respect to the supervised methods, our performance is comparable to the DRRN~\cite{taiimage} B1U9 setting, DRCN~\cite{kim2016deeply} and VSDR~\cite{kim2016accurate}; all three have networks with depth similar to ours.  The best results are reported by the DRRN B1U25~\cite{taiimage}, which has 50 convolution layers and is more than twice as deep as our network. 

Given the trend of the community to work on deep learning-based approaches for super-resolution, as well as the fact that no labelled data is required for training, one can work with (infintely) large datasets~\cite{Tong_2017_ICCV,lim2017enhanced}.  For fair comparison with state-of-the-art, however, we omit from this table the techniques which do not use the fixed 291 image training set as per the training protocol set by~\cite{taiimage,kim2016accurate}.  %We show in the last column some results with using just a subset of training data from Div2K~\cite{Agustsson_2017_CVPR_Workshops}, and already our PSNR increase up to $0.4$ db.  
% In the last column we provide the results of our algorithm in the Div2K data set~\cite{Agustsson_2017_CVPR_Workshops}~\footnote{for this experiment we follow the same experimental protocol but trained on Div2K data set.} where we work with patches of size $64\times 64$ sampled from the augmented dataset with a stride of $50$. The major of interest of this data set is that it has 800 images in the training set. 
% \subsection{Training Set Size}

 %Comparing the results of the kriging and the DKL we can see that kriging is improved when it is supervised. 

\subsection{Model uncertainty}
\begin{figure*}[tp!]
\begin{center}
\subfigure[Super resolution results of set 5 data set with scale $\times 3$.]{
\begin{tabular}{c}
\includegraphics[width=0.99\columnwidth]{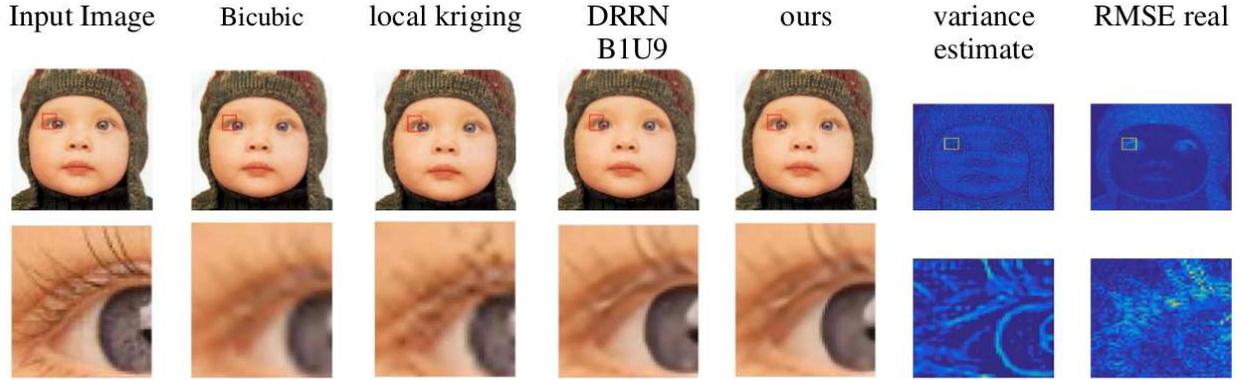}
\end{tabular}
	\label{fig:result_img1}
}

\subfigure[Super resolution results of set 5 data set with scale $\times 3$.]{
\begin{tabular}{c}
\includegraphics[width=0.99\columnwidth]{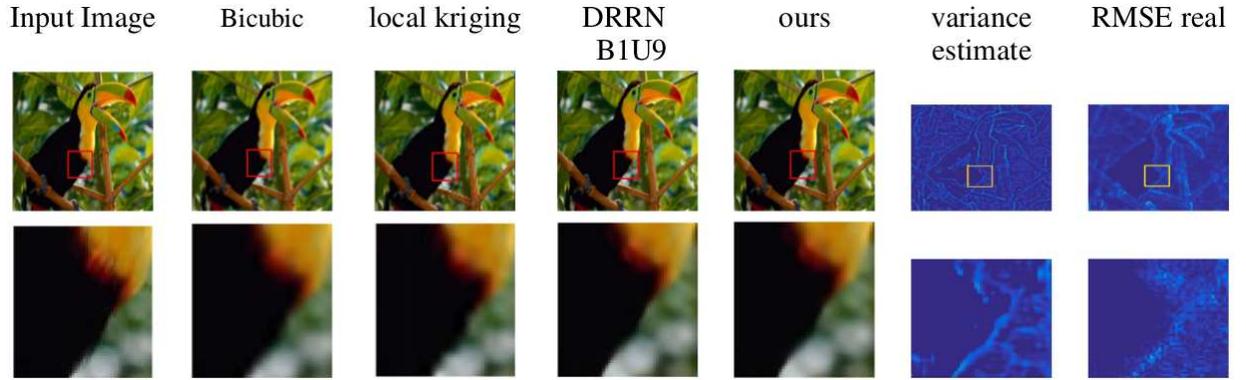}
\end{tabular}
	\label{fig:img3}
}
\end{center}
   \caption{Super resolution results }
\label{fig:result_img2}
\end{figure*}

One of the key strengths of our technique is the fact that we can estimate model uncertainty.  We show in Figure \ref{fig:result_img2} the  estimated variance for each pixel based on Equation \ref{Stat:form8}. To evaluate this equation, we need to apply covariance models for $f$ and $\tilde{f}$. We do this by first estimating the empirical covariance $ c(\tau) = \sum_i f(x_i)f(x_i+\tau)-\mu$ and then solving for $C_0$ and $\sigma$ of the Gaussian model in section 3.1 that is the closest to the empirical covariance.  
%One issue with this formula is that we need covariance depending on $f$ as a simplification we consider that $C(\tau) = \tilde{C}(\tau)$. %It happens that this approximation produce nice results.

%~\AY{you need to propose something resolution and or statement about why this is not such a problem, i.e. end the paragraph with a positive statement.  This was basically the open issue that the ICML reviewer also complained of no?}

The main advantage of our model of uncertainty is that we can have information on the black box CNN. It gives us feedback about the reliability of results provided by an unknown image.  One can see in Figure ~\ref{fig:result_img2} our uncertainty estimate. The estimated variance has a higher value than the real PSNR this is due to the fact that we have a noisy estimation of the PSNR from a low resolution image. Holistically, however, the are similar, in that areas with high variance corresponds to high PSNR. Quantitatively, we find for Set 5 that 91.9\% of the super-resolved pixel values fall within 3 standard deviations based on the estimated variance.
In addition, we evaluate the similarity between the images resulting from the PSNR and the one resulting from the variance in set 5. We use the correlation to measure the similarity and find that these images have high correlation up to 0.8.

\begin{table*}[tp!]
\begin{center}
\begin{tabular}{|c|c|c|c|c|c|c|c|c|c|}
\hline

\tiny{Dataset} & \tiny{Scale} & \tiny{Bicubic} & \tiny{local kringing} & \tiny{SRCNN} & \tiny{VDSR} & \tiny{DRCN} & \tiny{DRRN B1U9} & \tiny{DRRN B1U25} & \tiny{ours}\\
~ & ~ & \tiny{PSNR/SSIM} & \tiny{PSNR/SSIM} & \tiny{PSNR/SSIM} & \tiny{PSNR/SSIM} & \tiny{PSNR/SSIM} & \tiny{PSNR/SSIM} & \tiny{PSNR/SSIM}& \tiny{PSNR/SSIM}\\
\hline
\hline

\multirow{3}{*}{\tiny{Set 5}} & \scriptsize{$\times 2$}& \tiny{33.66/0.930}& \tiny{35.68/0.923} &\tiny{36.66/0.953}& \tiny{37.53/0.959} & \tiny{37.63/0.959} & \tiny{37.66/0.959} & \tiny{{37.74}/0.959} & \tiny{37.65/{0.960}}\\
& \scriptsize{$\times 3$} & \tiny{30.39/0.868} & \tiny{31.72/0.864}&\tiny{32.75/0.909} & \tiny{33.66/0.921} & \tiny{33.82/0.923} & \tiny{33.93/0.923} & \tiny{{34.04/0.924}} & \tiny{33.94/0.923}\\
& \scriptsize{$\times 4$} & \tiny{28.42/0.810}& \tiny{29.83/0.814} &\tiny{30.48/0.862} & \tiny{31.35/0.883} & \tiny{31.53/0.885} & \tiny{31.58/0.886} & \tiny{{31.68/0.889}} & \tiny{31.56/0.886}\\

\hline
\hline

\multirow{3}{*}{\tiny{Set 14}} & \scriptsize{$\times 2$}& \tiny{30.24/0.869} & \tiny{31.26/0.874} &\tiny{32.45/0.907}& \tiny{33.03/0.912} & \tiny{33.04/0.912} & \tiny{33.19/0.913} & \tiny{33.23/0.914} & \tiny{{33.10/0.912}}\\
& \scriptsize{$\times 3$} & \tiny{27.55/0.774}  & \tiny{27.94/0.781}&\tiny{29.30/0.822} & \tiny{29.77/0.831} & \tiny{29.76/0.831} & \tiny{29.94/0.831} & \tiny{{29.96}/0.834} & \tiny{29.81/{0.831}} \\
& \scriptsize{$\times 4$} & \tiny{26.00/0.703}  & \tiny{26.43/0.715} &\tiny{27.50/0.751} & \tiny{28.01/0.767} & \tiny{28.02/0.767} & \tiny{28.02/0.767} & \tiny{{28.18/0.770}} & \tiny{28.08/0.771}\\
\hline
\hline

\multirow{3}{*}{\tiny{B 100}} & \scriptsize{$\times 2$} & \tiny{29.56/0.843} & \tiny{31.06/0.848}&\tiny{31.36/0.888}& \tiny{31.90/0.8960} & \tiny{31.85/0.894} & \tiny{32.01/0.897} & \tiny{{32.05}/0.897} & \tiny{32.01/{0.896}}\\
& \scriptsize{$\times 3$} & \tiny{27.21/0.8431} & \tiny{27.86/0.7357}&\tiny{31.36/0.8879} & \tiny{28.82/0.7976} & \tiny{28.80/0.7963} & \tiny{28.91/0.7992} & \tiny{{28.95/0.800}} & \tiny{28.82/0.800}\\
& \scriptsize{$\times 4$} & \tiny{25.96/0.668} & \tiny{26.55/0.665}&\tiny{26.90/0.711} & \tiny{27.29/0.725} & \tiny{27.23/0.723} & \tiny{27.35/0.726} & \tiny{{27.38/0.728}} & \tiny{27.36/0.727}\\
\hline
\hline

\multirow{3}{*}{\tiny{Urban 100}} &  \scriptsize{$\times 2$} & \tiny{26.88/0.840} & \tiny{28.32/0.843}&\tiny{29.50/0.895}& \tiny{30.76/0.914} & \tiny{30.75/0.913} & \tiny{31.02/0.916} & \tiny{{31.23}/0.919} & \tiny{30.95/{0.921}}\\
& \scriptsize{$\times 3$}  & \tiny{24.46/0.735} & \tiny{25.06/0.715}&\tiny{29.50/0.799} & \tiny{27.14/0.715} & \tiny{27.15/0.828} & \tiny{27.38/0.833} & \tiny{{27.53/0.838}} & ~\\
& \scriptsize{$\times 4$}  & \tiny{23.14/0.658} & \tiny{23.70/0.627}&\tiny{24.52/0.722} & \tiny{25.19/0.752} & \tiny{25.14/0.751} & \tiny{25.35/0.757} & \tiny{{25.44/0.764}} & \tiny{25.10/0.749} \\
\hline
\end{tabular}
\end{center}
\caption{Results of various super resolution techniques. The column 3th and 4th are unsupervised. The other columns are trained on the 291 data set~\cite{yang2010image,MartinFTM01}. The Urban 100 set having no images at scale 3 we did not evaluate it.}
\label{table:results_Superresolution}
\end{table*}

%\subsection{Generalized kriging}
%
%We further explore two variants of our proposed deep kriging.  The first is to add a bias term $b$, so that we solve for 
%
%\begin{multline}
%\hat{f}^i_* = \mkern-36mu \sum_{k\in \{\|x-x^* \|_1 <\mathcal{K} \} }\mkern-36mu \omega_{k}(x^*) \tilde{f}_{k}^i  + b(x^*)\\
%\mbox{ with } \mkern-36mu \sum_{k\in \{\|x-x^* \|_1 <\mathcal{K} \}} \mkern-36mu \omega_{k}(x^*) = 1,
%\end{multline}
%where we learn both $\omega$ and $b$ with the same network as described in Section~\ref{sec:implementation}.  %in the same was as  are learned thanks to a network. 
%In this case, it improves our results by $0.3$ dB in PSNR in Urban 100 data set. However, such a solution deviates from the true ordinary kriging, in which case the statistical properties outlined in Section~\ref{sec:stats} no longer hold.
%%If we consider this kind of solution then our network will be quite comparable to the VSDR [ref] that learns a bias to add to the original image. This option should be considered if the subsampling process is not master.
%
%Alternatively, we also explore the possibility of stacking together consecutive kriging processes; however, this only increases the memory needs and does not improve the results significantly.  This is expected since the application of the kriging weights is linear, in which case stacked kriging processes will still collapse into a single linear model.

\section{Conclusions}
In this paper, we have proposed a joint deep learning and statistical framework for single image super-resolution based on a form of supervised kriging.  We solve for the kriging weights in the manner of a local dynamic filter and apply it directly to the low resolution image, all within a single network that can be learned end-to-end.  Since we work within the known statistical framework of kriging, we can estimate model uncertainty, something typically not possible for deep networks.  More specifically, we show through derivations that the statistical estimator generated by our network is unbiased and we calculate its variance.
%~\AY{X}.

\paragraph{Acknowledgement}
This research was funded by the German Research Foun-
dation (DFG) as part of the research training group GRK
1564 Imaging New Modalities.


\begin{thebibliography}{10}

\bibitem{chang2004super}
H.~Chang, D.~Yeung, and Y.~Xiong.
\newblock Super-resolution through neighbor embedding.
\newblock In {\em CVPR}, volume~1, pages I--I. IEEE, 2004.

\bibitem{keys1981cubic}
R.~Keys.
\newblock Cubic convolution interpolation for digital image processing.
\newblock {\em IEEE Trans. on Acoustics, Speech, and Signal Processing},
  29(6):1153--1160, 1981.

\bibitem{donoho2006compressed}
David~L Donoho.
\newblock Compressed sensing.
\newblock {\em IEEE Trans. on Information Theory}, 52(4):1289--1306, 2006.

\bibitem{yang2012coupled}
J.~Yang, Z.~Wang, Z.~Lin, S.~Cohen, and T.~Huang.
\newblock Coupled dictionary training for image super-resolution.
\newblock {\em IEEE Transactions on Image Processing (TIP)}, 21(8):3467--3478,
  2012.

\bibitem{lecun1998gradient}
Y.~LeCun, L.~Bottou, Y.~Bengio, and P.~Haffner.
\newblock Gradient-based learning applied to document recognition.
\newblock {\em Proceedings of the IEEE}, 86(11):2278--2324, 1998.

\bibitem{kim2016accurate}
J.~Kim, L.~Kwon Lee, and K.~Mu Lee.
\newblock Accurate image super-resolution using very deep convolutional
  networks.
\newblock In {\em CVPR}, 2016.

\bibitem{kim2016deeply}
J.~Kim, L.~Kwon Lee, and K.~Mu Lee.
\newblock Deeply-recursive convolutional network for image super-resolution.
\newblock In {\em CVPR}, 2016.

\bibitem{lim2017enhanced}
B.~Lim, S.~Son, H.~Kim, S.~Nah, and K.~M. Lee.
\newblock Enhanced deep residual networks for single image super-resolution.
\newblock In {\em CVPR Workshops}, 2017.

\bibitem{taiimage}
Y.~Tai, J.~Yang, and X.~Liu.
\newblock Image super-resolution via deep recursive residual network.
\newblock In {\em CVPR}, June 2017.

\bibitem{Sajjadi_2017_ICCV}
M.~S.~M. Sajjadi, B.~Scholkopf, and M.~Hirsch.
\newblock {EnhanceNet}: Single image super-resolution through automated texture
  synthesis.
\newblock In {\em ICCV}, Oct 2017.

\bibitem{MatheronBook}
Georges Matheron.
\newblock {\em Random sets and integral geometry}.
\newblock John Wiley \& Sons, 1975.

\bibitem{cressie2015statistics}
Noel Cressie.
\newblock {\em Statistics for spatial data}.
\newblock John Wiley \& Sons, 2015.

\bibitem{freeman2002example}
W.~Freeman, T.~Jones, and E.~Pasztor.
\newblock Example-based super-resolution.
\newblock {\em IEEE Computer Graphics and Applications}, 22(2):56--65, 2002.

\bibitem{peleg2014statistical}
T.~Peleg and M.~Elad.
\newblock A statistical prediction model based on sparse representations for
  single image super-resolution.
\newblock {\em IEEE Transactions on Image Processing (TIP)}, 23(6):2569--2582,
  2014.

\bibitem{dong2011image}
W.~Dong, L.~Zhang, G.~Shi, and X.~Wu.
\newblock Image deblurring and super-resolution by adaptive sparse domain
  selection and adaptive regularization.
\newblock {\em IEEE Transactions on Image Processing (TIP)}, 20(7):1838--1857,
  2011.

\bibitem{zhao2003wavelet}
S.~Zhao, H.~Han, and S.~Peng.
\newblock Wavelet-domain hmt-based image super-resolution.
\newblock In {\em International Conference on Image Processing (ICIP)},
  volume~2, pages II--953. IEEE, 2003.

\bibitem{zhang2012single}
K.~Zhang, X.~Gao, D.~Tao, and X.~Li.
\newblock Single image super-resolution with non-local means and steering
  kernel regression.
\newblock {\em IEEE Transactions on Image Processing (TIP)}, 21(11):4544--4556,
  2012.

\bibitem{anbarjafari2010image}
G.~Anbarjafari and H.~Demirel.
\newblock Image super resolution based on interpolation of wavelet domain high
  frequency subbands and the spatial domain input image.
\newblock {\em ETRI journal}, 32(3):390--394, 2010.

\bibitem{xu2010two}
L.~Xu and J.~Jia.
\newblock Two-phase kernel estimation for robust motion deblurring.
\newblock In {\em ECCV}, pages 157--170. Springer, 2010.

\bibitem{marquina2008image}
A.~Marquina and S.~Osher.
\newblock Image super-resolution by tv-regularization and bregman iteration.
\newblock {\em Journal of Scientific Computing}, 37(3):367--382, 2008.

\bibitem{he2011single}
H.~He and W.~C. Siu.
\newblock Single image super-resolution using {Gaussian} process regression.
\newblock In {\em CVPR}, 2011.

\bibitem{de2016dynamic}
B.~De Brabandere, X.~Jia, T.~Tuytelaars, and L.~Van Gool.
\newblock Dynamic filter networks.
\newblock In {\em NIPS}, 2016.

\bibitem{wilson2016deep}
A.~G. Wilson, Z.~Hu, R.~Salakhutdinov, and E.~P Xing.
\newblock Deep kernel learning.
\newblock In {\em AISTATS}, 2016.

\bibitem{damianou2013deep}
A.~Damianou and N.~Lawrence.
\newblock Deep gaussian processes.
\newblock In {\em Artificial Intelligence and Statistics}, pages 207--215,
  2013.

\bibitem{pronzato2017bayesian}
Luc Pronzato and Maria-Jo{\~a}o Rendas.
\newblock Bayesian local kriging.
\newblock {\em Technometrics}, pages 1--12, 2017.

\bibitem{meier2014local}
F.~Meier, P.~Hennig, and S.~Schaal.
\newblock Local gaussian regression.
\newblock {\em arXiv preprint arXiv:1402.0645}, 2014.

\bibitem{ioffe2015batch}
Sergey Ioffe and Christian Szegedy.
\newblock Batch normalization: Accelerating deep network training by reducing
  internal covariate shift.
\newblock In {\em International Conference on Machine Learning}, pages
  448--456, 2015.

\bibitem{yang2010image}
J.~Yang, J.~Wright, T.~Huang, and Y.~Ma.
\newblock Image super-resolution via sparse representation.
\newblock {\em IEEE Transactions on Image Processing (TIP)}, 19(11):2861--2873,
  2010.

\bibitem{MartinFTM01}
D.~Martin, C.~Fowlkes, D.~Tal, and J.~Malik.
\newblock A database of human segmented natural images and its application to
  evaluating segmentation algorithms and measuring ecological statistics.
\newblock In {\em ICCV}, volume~2, pages 416--423, July 2001.

\bibitem{bevilacqua2012low}
M.~Bevilacqua, A.~Roumy, C.~Guillemot, and M.~Alberi-Morel.
\newblock Low-complexity single-image super-resolution based on nonnegative
  neighbor embedding.
\newblock 2012.

\bibitem{huang2015single}
J.~Huang, A.~Singh, and N.~Ahuja.
\newblock Single image super-resolution from transformed self-exemplars.
\newblock In {\em CVPR}, pages 5197--5206, 2015.

\bibitem{wang2004image}
Z.~Wang, A.~Bovik, H.~Sheikh, and E.~Simoncelli.
\newblock Image quality assessment: from error visibility to structural
  similarity.
\newblock {\em IEEE Transactions on Image Processing (TIP)}, 13(4):600--612,
  2004.

\bibitem{Tong_2017_ICCV}
T.~Tong, G.~Li, X.~Liu, and Q.~Gao.
\newblock Image super-resolution using dense skip connections.
\newblock In {\em ICCV}, Oct 2017.

\end{thebibliography}
\end{document}